\documentclass[preprint]{elsarticle}
\usepackage{hyperref}
\usepackage{graphicx}
\usepackage{booktabs}
\usepackage{multirow}
\usepackage[caption=false]{subfig}
\usepackage{tikz}
\usetikzlibrary{positioning}
\usepackage{framed}
\usepackage{subfig}
\usepackage{comment}
\usepackage{epstopdf}
\usepackage{enumitem} 

\newcommand{\mm}[1]{{\fontfamily{cmtt}\selectfont #1}}

\journal{\parbox[t]{0.6\linewidth}{Simulation Modelling Practice and Theory\\ (\url{https://doi.org/10.1016/j.simpat.2020.102126}) }} 
 
\begin{document}

\begin{frontmatter}
\title{Quantitatively Assessing the Benefits of Model-driven Development in Agent-based Modeling and Simulation}

\author[udesc]{Fernando Santos\corref{corresponding}}
\cortext[corresponding]{Corresponding author}
\ead{fernando.santos@udesc.br}

\author[ufrgs]{Ingrid Nunes}
\ead{ingridnunes@inf.ufrgs.br}

\author[ufrgs]{Ana L. C. Bazzan}
\ead{bazzan@inf.ufrgs.br}

\address[udesc]{Universidade do Estado de Santa Catarina (UDESC), Ibirama, Brazil}
\address[ufrgs]{Universidade Federal do Rio Grande do Sul (UFRGS), Porto Alegre, Brazil}

\begin{abstract}  % put your abstract here!
The agent-based modeling and simulation (ABMS) paradigm has been used to analyze, reproduce, and predict phenomena related to many application areas. Although there are many agent-based platforms that support simulation development, they rely on programming languages that require extensive programming knowledge. Model-driven development (MDD) has been explored to facilitate simulation modeling, by means of high-level modeling languages that provide reusable building blocks that hide computational complexity, and code generation. However, there is still limited knowledge of how MDD approaches to ABMS contribute to increasing development productivity and quality. We thus in this paper present an empirical study that quantitatively compares the use of MDD and ABMS platforms mainly in terms of effort and developer mistakes. Our evaluation was performed using MDD4ABMS---an MDD approach with a core and extensions to two application areas, one of which developed for this study---and NetLogo, a widely used platform. The obtained results show that MDD4ABMS requires less effort to develop simulations with similar (sometimes better) design quality than NetLogo, giving evidence of the benefits that MDD can provide to ABMS.

\end{abstract}


\end{frontmatter}

%%%%%%%%%%%%%%%%%%%%%%%%%%%%%%%%%%%%%%%%%%%%%%%%%%%%%%%%%%%%%%%%%%%%%%%%%%%%%%%%%%%%%%%%%%%%%%%%%%%%%%%%%
%	INTRODUCTION
%%%%%%%%%%%%%%%%%%%%%%%%%%%%%%%%%%%%%%%%%%%%%%%%%%%%%%%%%%%%%%%%%%%%%%%%%%%%%%%%%%%%%%%%%%%%%%%%%%%%%%%%%
\section{Introduction}\label{sec:introduction}

Agent-based simulations have been widely used to understand the emergent behavior of complex systems. These systems are composed of multiple entities, or agents, which can interact with each other and are situated in an environment that they can perceive and modify through their actions.  Building agent-based simulations is a challenging task that has been widely investigated in the context of \emph{agent-based modeling and simulation} (ABMS), a simulation paradigm that uses autonomous agents and multiagent systems to reproduce and explore a phenomenon under investigation. The ABMS paradigm has been used in many application areas, such as traffic, ecology, economics, and epidemiology~\cite{Macal&North2014}. According to \citet{Macal&North2014}, ABMS is selected as the simulation paradigm in such areas because it can explicitly incorporate the complexity arising from individual behavior and interactions that exist in real settings. Agents can furthermore be endowed with learning or evolutionary capabilities to adapt to changes in themselves or the environment.  Artificial intelligence techniques that provide such capabilities are well established and can be incorporated into agents leading to more realistic simulations~\cite{Kluegl&Bazzan2012}. 

Many alternatives have been proposed for easing the development of agent-based simulations. Platforms, such as NetLogo~\cite{Wilensky1999} and Repast~\cite{North+2006}, have been widely used because they provide programming environments that offer language constructs to explicitly represent agents and their interactions, the environment as well as the creation and initialization of entities and agents. These platforms, however, demand previous expertise in ABMS and programming. Researchers have already argued about the importance of providing alternatives for modeling and creating agent-based  simulations by means of building blocks at a higher abstraction level~\cite{Hamill2010,Kluegl&Bazzan2012,Wellman2016}. Such approaches would potentially decrease the development time and effort to build agent-based simulations.

Approaches towards this direction have been built upon the fundamentals of \emph{model-driven development} (MDD)~\cite{Atkinson&Kuhne2003,Schmidt2006}, a software development approach in which models are considered \emph{first-class citizens} and the development is driven by modeling artifacts~\cite{Stahl+2006}. These models are used to \mbox{(semi-)}automatically generate the source code of software systems and thus play a leading role in the development process. Previous work on the use of MDD in industry showed that MDD approaches can increase productivity by a 5--10x factor~\cite{Tolvanen&Kelly2016} and that the productivity increases because the modeling effort is focused on \emph{domain concerns} instead of programming statements~\cite{Sprinkle+2009}. 

An MDD approach is composed of a domain-specific language (DSL) and model transformations for code generation~\cite{Schmidt2006}. The former allows effectively expressing domain concepts, while the latter introduces automation. A DSL is a modeling or programming language designed for a particular domain that trades generality for expressiveness. DSLs provide high-level, off-the-shelf, abstractions for business-related concepts and processes. By doing so, MDD approaches raise the abstraction level of models, allowing developers to refer to the desired functionality of the system instead of the details related to its development and deployment~\cite{Beydeda2005}. As noted by \citet{Stahl+2006}, abstraction in MDD models does not mean vagueness, but compactness and reduction to the essence. 

Despite qualitative arguments that motivate the use of MDD in ABMS, there is a need for concrete evidence that it in fact promotes its assumed benefits, i.e.\ by raising the level of ABMS by means of domain-specific abstractions, there is an increase in productivity. Previous studies provide examples of use~\cite{Garro+2013,Ghorbani+2014,GomezSanz+2010,Kluegl&Davidsson2013,Ozik+2015}, evaluation of  comprehension~\cite{Santos+2017compsac} and comparison of software size as an estimation of effort~\cite{Santos+2018simpat}. Nevertheless, there is a lack of an assessment of the development effort and quality involving development tasks performed by simulation developers. 
 
In response, we in this paper present an empirical study to quantitatively assess the benefits promoted by MDD in ABMS. More specifically, we take an existing approach in this context, namely MDD4ABMS~\cite{Santos+2017aamas,Santos+2017compsac}, and compare it with NetLogo, a widely used agent-based simulation platform. We selected MDD4ABMS because it not only provides common ABMS concepts as part of its graphical domain-specific language but also has an extension that includes concepts specific to an application area~\cite{Santos+2018simpat}, which is \emph{adaptive traffic signal control}. This approach thus effectively exploits the trade-off between expressivity and generality that is made in MDD~\cite{Stahl+2006}. In order to generalize our results to other domains, we provide a novel extension of MDD4ABMS, focusing on an alternative application area, the \emph{spread of diseases}. Our empirical study consists of a comparison between MDD4ABMS and NetLogo when they are used by developers to build agent-based simulations in these two application areas. We collected data associated with the design quality of developed simulations and time spent to develop them. We also made a subjective evaluation of the two development approaches by means of questionnaire answered by participants. The obtained results show that MDD4ABMS requires less effort to develop simulations with similar (sometimes better) design quality than NetLogo, giving evidence of the benefits that MDD can provide to ABMS.

Our contributions thus are: (i) an extension of the MDD4ABMS approach, targeting the application area of spread of diseases; (ii) a study protocol to evaluate MDD approaches for ABMS; and (iii) an empirical study that compares MDD4ABMS and NetLogo in two application areas. 

%%%%%%%%%%%%%%%%%%%%%%%%%%%%%%%%%%%%%%%%%%%%%%%%%%%%%%%%%%%%%%%%%%%%%%%%%%%%%%%%%%%%%%%%%%%%%%%%%%%%%%%%%
%	RELATED WORK
%%%%%%%%%%%%%%%%%%%%%%%%%%%%%%%%%%%%%%%%%%%%%%%%%%%%%%%%%%%%%%%%%%%%%%%%%%%%%%%%%%%%%%%%%%%%%%%%%%%%%%%%%

\section{Related Work}\label{sec:related-work}

Previous work on the use of MDD showed that it brings benefits to the development of general software systems in different domains, such as automotive manufacturing, mobile devices and internet of things, telecommunications, and military~\cite{Sprinkle+2009,Torchiano+2013,Tolvanen&Kelly2018}. \citet{Tolvanen&Kelly2018} investigated the development and use of MDD approaches in a variety of domains and concluded that the investment required to create complete MDD solutions is modest when appropriated toolkits are used.  The effort required to build these solutions is paid back quickly, providing a good return of investment~\cite{Tolvanen&Kelly2018}.  Previously,~\citet{Mohagheghi+2013} have already identified that using the appropriate toolkits for creating an MDD solution is crucial for the industrial adoption of MDD. In a study with three industrial cases, they reported that merging different mainstream toolkits in a seamless MDD solution demanded several transformations and increased the required implementation effort and complexity.

Lately, work has been done to apply or evaluate MDD approaches in domains such as embedded software~\cite{Akdur+2018}, game development~\cite{Tang&Hanneghan2011,Zhu&Wang2019}, and even to automate the migration of legacy models of software components~\cite{Schuts+2018}.  In a survey with MDD practitioners from different industrial sectors,~\citet{Wittle+2013models} reported that some form of MDD is practiced widely across a diverse range of industries. Companies who successfully applied model-driven techniques did so by using (or even creating) languages specifically developed for their domains rather than using general purpose languages such as the unified modeling language (UML).  \citet{Wittle+2013models} also noticed that companies that target a particular domain are more likely to use model-driven techniques than companies that develop generic software. Previous work has already shown that the more specific the modeling language, the higher the chance of success of an MDD approach~\cite{Hutchinson+2011}.

MDD has already been considered by the modeling and simulation community as a viable approach for producing executable simulations from models. Models specified using either the Business Process Model and Notation (BPMN) or the DEVS Modeling Language (DEVSML) were considered by \c Cetinkaya and coleagues~\cite{Cetinkaya+2013bpmn,Cetinkaya+2013devsml}. In their work, model transformations were proposed for generating simulations targeted to the Discrete Event Systems (DEVS) paradigm.  Bocciarelli and D'Ambrogio also considered BPMN models~\cite{Bocciarelli+2012}, in addition to models specified using the Systems Modelling Language (SysML)~\cite{Bocciarelli&DAmbrogio2014}, but proposed transformations for generating simulations targeted the Distributed Simulation (DS) paradigm. In contrast, the MDD4ABMS approach targets the ABMS paradigm, in which simulations are specified considering agents, interactions, and the environment. Modeling agent systems using languages conceived for modeling systems in paradigms other than ABMS would potentially raise expressiveness issues, similarly to what was previously reported with respect to using UML for modeling agent-systems~\cite{Bauer&Odell2005}. The MDD4ABMS modeling language, instead, provides elements for modeling recurrent aspects of agent-based simulations, such as flow control and learning capabilities, which reduce the abstraction gap and thus improve expressiveness and productivity.

In the remaining of this section, we discuss work related to MDD and simulations. We start presenting existing MDD approaches targeted to multiagent systems, and then we narrow the review to approaches targeted to ABMS. We conclude this section with a discussion on the effectiveness evaluation in these related approaches.

\subsection{MDD and Multiagent Systems}
With respect to model-driven development of agent and multiagent systems in general,  \citet{Bauer&Odell2005} investigated the use of UML for modeling agent-systems. In spite of being a modeling language widely adopted in industry to specify software systems, the authors concluded that UML---and even its extension Agent UML~\cite{Bauer+2001}--- is not expressive enough to specify intricate aspects of agent-based systems. For example, off-the-shelf constructs to express aspects such as reproduction and emergent phenomena are missing. According to the authors, a useful way of representing such aspects, at a higher level of abstraction, should be developed so as to model agent systems effectively.
\citet{Celaya+2007} and \citet{Marzougui+2010} considered modeling multiagent systems via Petri nets. Although they showed the value of Petri nets for model checking, these approaches rely on abstract agent architectures that cover only basic aspects of multiagent systems. As with UML-based approaches, complex aspects should be specified from scratch as well. 

\citet{Kardas2013} reviewed a selection of model-driven approaches and methodologies for multiagent systems. 
Although there are MDD alternatives that provide support for modeling and code generation, the author noticed that in most situations, such code is generated only at the template level, and a significant amount of code needs to be manually completed. Therefore, feasibility and effectiveness of these approaches are limited because the amount and quality of the automatically generated multiagent system components appear to be insufficient.  Furthermore, it is important to notice that these alternatives are focused on multiagent models, and thus simulation aspects are uncovered.

\subsection{MDD and ABMS}
MDD has already been considered for the development of agent-based simulations. There are MDD approaches that rely on the unified modeling language (UML) for specifying simulation models~\cite{Duarte&DeLara2009,Garro+2013,Iba+2004}. However, as reported by \citet{Bauer&Odell2005}, UML does not provide expressive constructs to specify high-level aspects of agent-based simulations, which often should be specified from scratch thus compromising the effectiveness of these UML-based MDD approaches. MDD approaches that do not rely on UML for modeling range from extensions of existing agent methodologies to new metamodels with modeling languages and model transformations, discussed next.

Among the approaches focused on agent-based simulations, the AMASON~\cite{Kluegl&Davidsson2013} metamodel covers only basic structures and dynamics of agent-based simulations. The MAIA~\cite{Ghorbani+2013} metamodel captures social concepts such as norms and roles. \citet{Ribino+2014} proposed a conceptual metamodel to be used as a guideline and concept repository for designing simulations. \citet{Hahn+2009} introduced PIM4Agents, a platform-independent metamodel divided into viewpoints that capture agent, organizational, interaction, role, behavioral, and environmental aspects. In the IODA methodology~\cite{Kubera+2011}, behaviors are encoded as an interaction matrix, which can be seen as a DSL for specifying the simulation dynamics. Overall, these metamodels support only abstract agent-based simulation concepts, leaving much left to be developed in specific applications. The MDD4ABMS approach~\cite{Santos+2017aamas}, target of the present study, provides support to additional agent-based simulation concepts, such as recurrent environment topologies and agent capabilities (e.g., learning). 

Although these contributions arguably provide many benefits, studies that evaluate the concrete benefits of existing proposals while being used by developers or other potential users are still missing. In order to show that MDD approaches are \emph{effective} by improving productivity or increasing quality, we must go beyond showing that it is \emph{feasible} to be used in practice. With respect to related work, we observed that they limit themselves to showing examples and case studies that demonstrate the \emph{feasibility} of their MDD approaches---the ability to model and, in some cases, to generate code. Although methodologies for evaluating MDD approaches point out user studies as the method for evaluating effectiveness~\cite{Mohagheghi+2009,Kahraman&Bilgen2015,Challenger+2015}, no evaluation with humans involving development tasks was reported in related work. 

In summary, despite being able to generate code from models, there is no evidence that the effort required to create models in these MDD approaches is lower than developing the simulation directly on the target simulation environment. With our study, we thus filled this gap in, assessing whether MDD can indeed improve the development of agent-based simulations.

%%%%%%%%%%%%%%%%%%%%%%%%%%%%%%%%%%%%%%%%%%%%%%%%%%%%%%%%%%%%%%%%%%%%%%%%%%%%%%%%%%%%%%%%%%%%%%%%%%%%%%%%%
%	MDD4ABMS (the contribution)
%%%%%%%%%%%%%%%%%%%%%%%%%%%%%%%%%%%%%%%%%%%%%%%%%%%%%%%%%%%%%%%%%%%%%%%%%%%%%%%%%%%%%%%%%%%%%%%%%%%%%%%%%

\section{MDD4ABMS and Extensions}\label{sec:mdd4abms}

MDD4ABMS~\cite{Santos+2017aamas,Santos+2017compsac} is a model-driven approach that supports the specification of agent-based simulation aspects.  MDD4ABMS is built upon a core metamodel that abstracts recurrent concepts used when developing agent-based simulations. Extensions to the core metamodel are available to incorporate additional, domain-specific, aspects. The MDD4ABMS approach also provides both a domain-specific modeling language and a modeling tool. 

An amount of effort was required to build the MDD4ABMS approach and extend it to incorporate additional domain-related aspects. As reported in software engineering literature, the provision of high-level domain-related abstractions for specifying models is one of the pillars of MDD~\cite{Atkinson&Kuhne2003, Schmidt2006}. The effort to build an MDD approach that provides such abstractions is paid back~\cite{Tolvanen&Kelly2018}, given that with such an MDD approach, developers can focus on domain concepts and this leads to productivity gains~\cite{Sprinkle+2009,Tolvanen&Kelly2016}. This paper aims at verifying whether this is the case for MDD4ABMS. 

We next describe the MDD4ABMS core metamodel. The metamodel extensions that provide support for adaptive traffic signal control and spread of disease simulations are presented afterwards.

\subsection{Core Metamodel}\label{sec:core-metamodel}
The core metamodel abstracts basic, recurrent concepts found in agent-based simulations~\cite{Santos+2017aamas,Santos+2017compsac}. These concepts include topologies for the simulated environment, in particular grids, Cartesian spaces, and graphs.

Additionally, the core metamodel also abstracts entities and agents with attributes. An \mm{Entity}\footnote{Monospaced typeface denotes elements included in the diagram presented later in Figure\ref{fig:abmsmm-aamas2019-capabilities-disease-statemachine}} is a relevant object that exists in a simulation. An \mm{Agent} is a particular kind of entity that may have \emph{capabilities}. A capability is an agent ability frequently used in simulations (e.g., mobility, life cycle). In the core metamodel, each of such capabilities is abstracted as an \mm{AgentCapability} element. The creation and initialization of both agents and the environment are represented as \emph{creational strategies}. Strategies for creating those elements according to popular file formats are provided, such as geographic information system (GIS) and open street map (OSM) files.

To mitigate scalability issues, the core metamodel introduces the notion of \emph{concern} for partitioning models, allowing a clear separation of the concepts associated with different aspects of the simulation. The creational strategies, agent capabilities, and the ability to clearly separate concerns distinguish MDD4ABMS from other classic approaches and platforms for agent-based simulation. Our previous study showed the benefits of MDD4ABMS for specifying and understanding simulations~\cite{Santos+2017compsac}. 

In addition to the agent-based simulation metamodel, the MDD4ABMS approach also provides both a domain-specific modeling language, named \emph{ABStractLang}, and a modeling tool, named \emph{ABStractme}~\cite{Moreira+2017}. AbstractLang is a visual modeling language that provides building blocks for specifying simulations according to the MDD4ABMS metamodel. The \emph{ABStractme} modeling tool is an Eclipse\footnote{\url{https://www.eclipse.org/}} plugin that allows the creation of \emph{ABStractLang} diagrams. \emph{ABStractme} also provides automatic code generation for the NetLogo simulation platform. 

\subsection{Adaptive Traffic Signal Control}\label{sec:atsc-metamodel}
The core metamodel has an extension that provides support for adaptive traffic signal control simulations~\cite{Santos+2018simpat}. In this application area, agents are in charge of managing traffic signal controls at intersections of a traffic network. The goal in these simulations is to evaluate whether agents can autonomously improve the flow of vehicles by using sophisticated decision-making strategies such as learning. Agents have to consider concepts from the traffic control domain, such as traffic signal phases (periods of time in which a subset of the traffic lights are set to green to allow the traffic flow in a particular direction) and plans (a set of phases plus the sequence in which they are activated). 

The ability to manage the flow in a set of streams (e.g., traffic lanes) is abstracted as a \emph{flow control} agent capability. To specify policies for managing such a flow, the following decision-making strategies frequently used in this application area were incorporated into the extended metamodel as agent capabilities: (i) \emph{state machine} agent capability, to represent a fixed decision policy by means of a \mm{State Machine}, which is composed of \mm{States}, \mm{Transitions}, and \mm{Triggers}; (ii) \emph{adaptation} agent capability, to represent an adaptive decision policy by means of an adaptation criterion; and (iii) \emph{reinforcement learning} agent capability, to allow an agent to learn a decision policy through experience.

The adaptive traffic signal control extension was previously evaluated considering software size as an estimation of effort~\cite{Santos+2018simpat}. However, no MDD4ABMS evaluation of effort with humans was considered so far, and therefore this is one main novel contribution in the present paper. Given that it was extended to a single application area, we introduce in this paper a proposed extension of MDD4ABMS targeting the spread of diseases, allowing us to evaluate the use of MDD4ABMS in two application areas. We next describe the new MDD4ABMS extension.

\subsection{Spread of Diseases}\label{sec:disease-metamodel}

\citet{Isern&Moreno2015} reported agent-based simulations as one trend in health informatics, which served as motivation for us to extend MDD4ABMS to the spread of diseases application area and use it in our empirical study. To extend the MDD4ABMS metamodel, we followed a bottom-up approach. We selected existing agent-based simulations of spread of diseases, and  conducted a domain analysis activity to identify recurrent concepts adopted in that simulations. In addition to the simulated environment and agents---concepts that were already supported by MDD4ABMS---the domain analysis activity revealed that these simulations adopt the compartmental model of Kermack~\&~McKendrick~\cite{Kermack&McKendrick1932} for specifying the spread of the disease.  

\paragraph{Compartmental Model}

In the compartmental model of Kermack~\&~McKendrick, individuals within a population are categorized into compartments. For example, the simplest model, called SIR, adopts the following three compartments: \textbf{S}usceptible, for individuals not exposed to the disease; \textbf{I}nfected, for individuals with the disease; and \textbf{R}ecovered for individuals that have successfully cleared the disease. The dynamics of the disease is represented by transitions between these compartments, which are governed by rates. A transmission rate specifies the probability of the disease being passed between agents when they interact, and a recovery rate specifies the rate at which infected agents recover from the infection and therefore determines the disease duration. Deaths caused by the disease can be incorporated via a death rate, and temporary immunity is obtained by specifying a duration for the \textbf{R} compartment. 
There are extensions to the SIR model that consider additional compartments to represent diseases with particular characteristics. A \textbf{P}assive Immunity compartment can be considered if individuals could have temporary immunity to a disease, leading to the so-called PSIR model. For diseases with an incubation period, an \textbf{E}xposed compartment can be incorporated to categorize those individuals that, although infected, yet do not manifest symptoms, leading to the SEIR model.
Finally, to run a simulation, it is mandatory to specify details of how the disease is introduced in the population of agents, such as the number of infected agents at the beginning of the simulation and whether the introduction is periodic. 

\begin{figure}
	\centering
	\includegraphics[angle=90,origin=c,width=.99\linewidth]{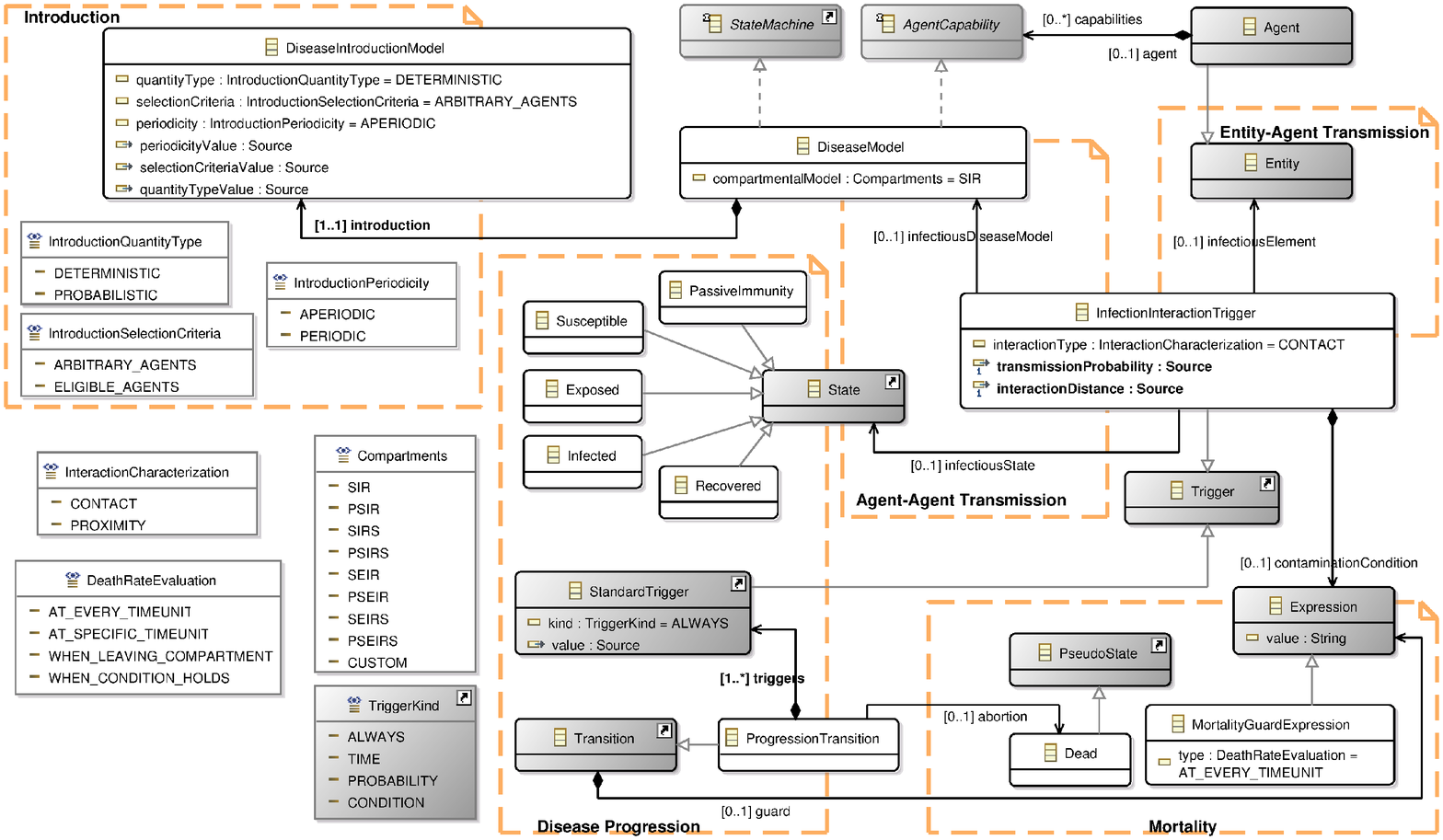}
	\caption{Spread of disease metamodel. Elements from the original metamodel are in gray.}
	\label{fig:abmsmm-aamas2019-capabilities-disease-statemachine}
\end{figure}

% Compartmental model as a state machine
In the extended MDD4ABMS, the compartmental model is abstracted as a state machine: compartments are states of the state machine, and transitions between compartments are transitions between states. Figure~\ref{fig:abmsmm-aamas2019-capabilities-disease-statemachine} shows the extended metamodel with a \mm{DiseaseModel} state machine that specializes \mm{StateMachine}. Elements from the original metamodel are highlighted with gray. Although the MDD4ABMS metamodel already specifies state machine elements (previously described in Section~\ref{sec:atsc-metamodel}), some of these elements are specialized to incorporate additional semantics. Each agent that is subject to a disease is in charge of managing its compartmental model and, therefore, the disease state machine is also an \mm{AgentCapability} (previously described in Section~\ref{sec:core-metamodel}). The compartmental models identified during the domain analysis are enumerated in the \mm{Compartments} element. One of these models is assigned to the state machine to specify which compartmental model is adopted and therefore of which states the machine is composed. Specializations of the \mm{State} element are specified to represent the identified compartments.  Compartments other than those specified in the metamodel are supported by the \mm{custom} compartmental model, which enables the instantiation of customized states and transitions.

\begin{figure}[t!]
	\centering
	\includegraphics[width=.91\linewidth]{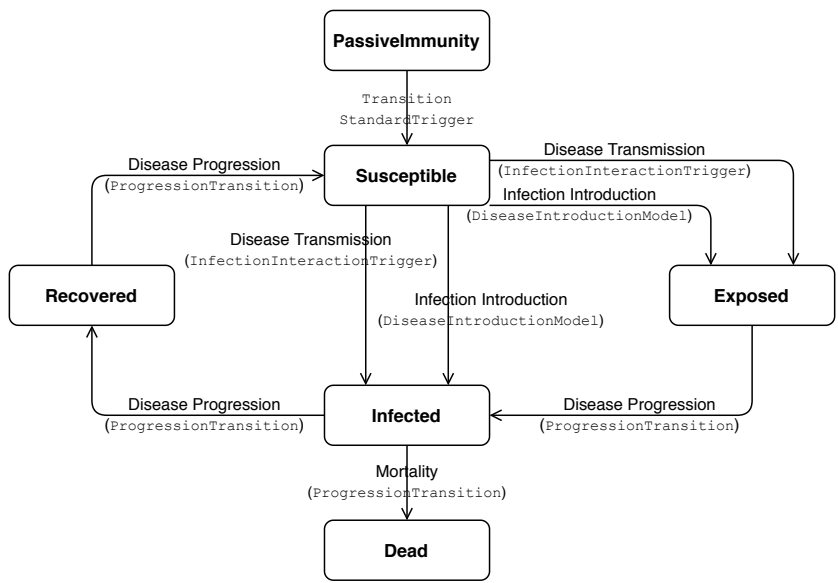}
	\caption{Behavioral view of the disease model state machine.}
	\label{fig:abmsmm-simpat2019-disease-statemachine-transitions}
\end{figure}

The behavioral view of the disease model state machine is specified in the state transition diagram shown in Figure~\ref{fig:abmsmm-simpat2019-disease-statemachine-transitions}. An agent subject to a disease has its state changed according to the transitions between states. The metamodel element(s) that specify these transitions are shown with \texttt{monospaced} font. The specified transitions are related to the \emph{disease transmission}, \emph{disease progression}, \emph{mortality}, and \emph{infection introduction}. These aspects are detailed as follows. 

\paragraph{Disease Transmission}

Disease transmission may take place whenever agents interact. From the domain analysis we identified that such an interaction is characterized either by physical contact or by spatial proximity. Therefore, from the spread of disease perspective, interaction between two agents occurs when the distance between them is below a given threshold, whose value depends the disease. The disease transmission is specified as transitions from the susceptible state to either the exposed or infected state. The \mm{InteractionCharacterization} element enumerates the two types of transmission interaction identified during the domain analysis: \emph{proximity}, for transmissions that occurs at a given distance; and \emph{contact}, for transmissions that occurs only when the distance between agents is zero. The customized trigger \mm{InfectionInteractionTrigger} models transmission interactions and specifies both the transmission probability and the interaction distance as \mm{Source} elements.\footnote{A source element abstracts the provision of values~\cite{Santos+2017compsac}.} For infections caused by a disease transmission during inter-agent interactions, the trigger refers to the disease model of the other agent involved in the interaction and its infectious state. For infections caused by interactions with contaminated objects (e.g., contaminated water), the trigger refers to the infectious entity. In both cases, a contamination condition expression may be specified to determine whether the transmission will indeed occur (e.g., bacterial concentration level in water below a particular threshold). 

\paragraph{Disease Progression}

After being infected, the current compartment of the agent changes according the disease progression. The compartmental model represents this progression as transitions governed by rates. These rates are related to the duration of each disease stage. In our metamodel, the duration of a particular compartment is specified by triggers associated with its state outgoing transitions. Each of these transitions represents, therefore, a progression to the next compartment. Additional semantics is incorporated by the  specialized state machine transition \mm{ProgressionTransition}. The following four distinct ways of specifying the duration of a particular compartment were identified during the domain analysis and are abstracted in the metamodel:  \emph{probabilistic}, in which a rate is specified and the transition is triggered probabilistically; \emph{deterministic}, in which a fixed period of time is specified and the agent stays in the compartment for that period; \emph{conditional}, in which a condition is specified and the agent stays in the compartment until this condition is met; and \emph{custom}, which allows combining previous durations. These four specifications use the trigger kinds enumerated in the \mm{TriggerKind} element. 

\paragraph{Mortality}

Deaths caused by the disease are specified as transitions to an additional pseudo-state called \mm{Dead}. Once the disease state machine reaches this pseudo-state, the agent dies. The domain analysis revealed that the circumstance under which death rates are evaluated varies from simulation to simulation. The following circumstances, enumerated in the \mm{DeathRateEvaluation} element, were identified: \emph{at every timeunit} and \emph{specific timeunit} define that the death rate is evaluated at every timeunit or at a particular timeunit, respectively; \emph{when condition holds} defines that the death rate is evaluated whenever a condition holds (e.g., when the agent runs out of energy); and \emph{when leaving compartment} defines that the death rate is evaluated only when the compartment duration has elapsed and the state machine is moving to the next compartment. The first three circumstances are represented in the metamodel as a guard expression (\mm{MortalityGuardExpression}) associated with the transition to the \mm{Dead} pseudo-state, which also has a probabilistic trigger whose value is the death rate. For the last circumstance, \emph{when leaving compartment}, there must be an outgoing progression transition departing from the state to which the death rate is applied. After being triggered, which means that the disease state machine is leaving the compartment, this transition may be aborted due to the death rate. If so, the state machine moves to an \mm{abortion} state, which is the \mm{Dead} state. 

\paragraph{Infection Introduction}

The metamodel also supports the specification of how the infection is introduced in the population. The domain analysis revealed that the infection introduction is governed by the following aspects. 

\begin{itemize}
\item \emph{Quantity}: of how many entities or agents the infection is introduced. Quantity is either \textit{deterministic} or \emph{probabilistic}. A deterministic quantity specifies the number of entities/agents that will be infected, while a probabilistic quantity specifies the chance that any entity/agent has to be infected.
\item \emph{Selection criterion}: determines which entities are considered for having the infection introduced. Selection criterion is either \emph{arbitrary} or \emph{eligible}. Arbitrary means that any entity/agent can be infected, while eligible means that only entities that meet the eligibility criterion are considered for being infected. In both cases, only susceptible agents are considered. 
\item \emph{Periodicity}: in which the infection is introduced. Periodicity is either \emph{aperiodic} or \emph{periodic}. In an aperiodic introduction, infection is introduced at the beginning of the simulation, while a periodic introduction (re)introduces the infection periodically. 
\end{itemize}

These aspects are represented in the metamodel by the \mm{IntroductionQuantityType}, \mm{IntroductionSelectionCriteria}, and \mm{IntroductionPeriodicity}, respectively. Note that infection introduction is beyond the agent scope: it is not the agent that decides whether it was selected for being infected. If it were, the agent would have to know which others were also selected in order to respect the total number of infected agents; however, an agent does not have this global knowledge. Therefore, infection introduction is a task executed by the simulation controller. Consequently, all the aspects that govern infection introductions are specified as \mm{DiseaseIntroductionModel} elements, which are read by the controller during the simulation execution.

% Minor extensions
Minor extensions to the MDD4ABMS core allow the specification of additional elements required to run and analyze spread of disease simulations. A \emph{mobility agent capability} is used to abstract the way agents move around the environment. Additionally, an \emph{output dataset} element allows specifying the data to be collected for further analysis of the simulation results. Lastly, both the metamodel and modeling language have now an abstraction of external agent capabilities that are provided as source code libraries, so as to allow the designer to incorporate behaviors not covered by the MDD4ABMS approach yet.

\paragraph{Code Generation}

Model-to-code transformations were developed to produce ready-to-run code for the NetLogo simulation platform. The transformation rules are specified and documented using the Xpand\footnote{\url{http://www.eclipse.org/modeling/m2t/?project=xpand}} template language. Each Xpand template describes the source code statements that are generated for each spread of disease metamodel element. To guarantee that generated simulations are adequately coded and correctly implement the expected behaviors, a verification procedure was performed, as recommended by~\citet{Crooks&Hailegiorgis2014} and \citet{Iba+2004}. Such a verification consisted of detailed inspections of the source code and debugging sessions, to ensure that all the units of code are performing their corresponding operations and are correctly integrated to implement each agent capability. Additionally, unit tests were executed during the development of the transformation rules to assert they are producing the expected NetLogo source code for different parts of our metamodel. We recall that these unit tests were conducted during the development of the source code generator, and therefore are neither related nor used in the empirical evaluation described in this paper. Finally, existing NetLogo simulations were specified using the metamodel elements, and the generated simulation code was executed to validate the produced outputs.

\begin{figure}[b!]
	\centering
	\includegraphics[width=1\linewidth]{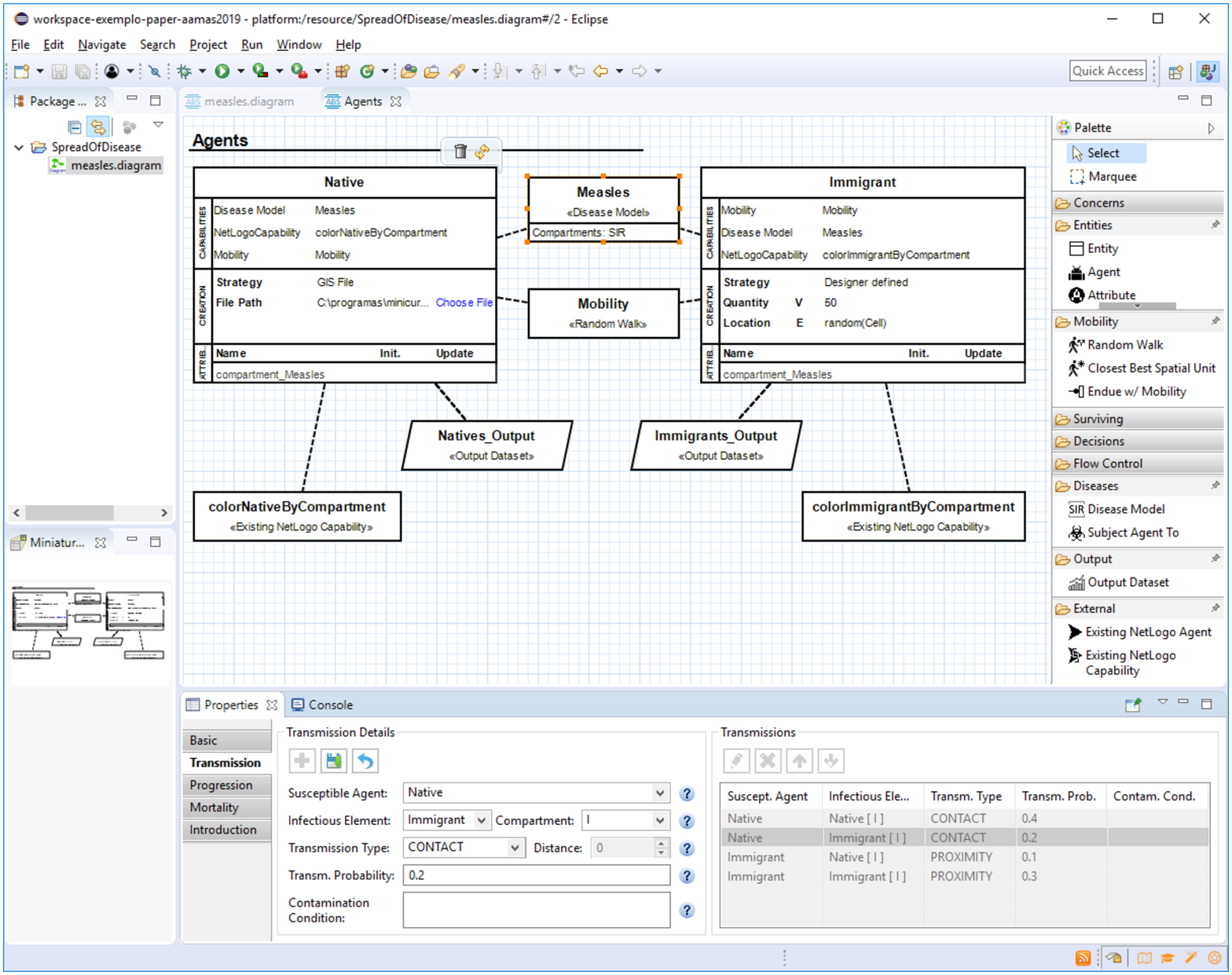}
	\caption{ABStractLang in the ABStractme tool.}
	\label{fig:abstractme}
\end{figure}

\paragraph{Concrete syntax}

Finally, the concrete syntax of the ABStractLang language is also extended to provide elements and views for modeling such a disease model.  As all other existing agent capabilities, the disease model agent capability is represented as a box that shows the disease name and the selected compartmental model. This visual element can then be connected to the agents that are subject to the disease, fostering the reuse of the disease specification. Figure~\ref{fig:abstractme} illustrates the concrete syntax of the ABStractLang language with an example model that specifies two types of agent (\emph{native} and \emph{immigrant}), both subject to the \emph{measless} disease model. The example is shown within the interface of the ABStractme tool~\cite{Moreira+2017}, which includes: a diagram editor on the center, in which the designer specifies the simulation elements using the ABStractLang language; a properties section on the bottom that shows the ABStractLang view related to the model element that is currently selected in the diagram; and a palette of elements on the right from which the designer can drag and drop them into the diagram.

%%%%%%%%%%%%%%%%%%%%%%%%%%%%%%%%%%%%%%%%%%%%%%%%%%%%%%%%%%%%%%%%%%%%%%%%%%%%%%%%%%%%%%%%%%%%%%%%%%%%%%%%%
%	USER STUDY
%%%%%%%%%%%%%%%%%%%%%%%%%%%%%%%%%%%%%%%%%%%%%%%%%%%%%%%%%%%%%%%%%%%%%%%%%%%%%%%%%%%%%%%%%%%%%%%%%%%%%%%%%

\section{Study Settings}\label{sec:user-study}

Given that we introduced the main object of study of our empirical evaluation (existing and novel extensions), we now detail our study design and its participants to compare MDD4ABMS and Netlogo.

\subsection{Goal and Research Questions}

To design our study, we followed the principles of experimental software engineering, using the \emph{goal-question-metric} (GQM) paradigm~\cite{Basili+1986}. Our study goal, based on the GQM template, is as follows.

\begin{framed}
\noindent \textbf{Study Goal:} \emph{to assess the benefits of using an MDD approach to develop agent-based simulations, evaluate MDD4ABMS in comparison with NetLogo from the perspective of the researcher, as they are used by developers with little expertise in ABMS in a multi-project study}.
\end{framed}

The NetLogo platform was chosen as the baseline for evaluating the benefits provided by MDD4ABMS based on its popularity and because it provides high-level commands for ordinary operations frequently required in agent-based simulations, such as moving agents or finding elements located within a given radius. These advantages of NetLogo over other simulation platforms have already been reported in literature~\cite{Railsback+2006}. 
Additionally, in our study, participants run simulations to verify the developed features. Thus, NetLogo was chosen given that it is an approach in which simulations can be successfully developed and ran by participants. As noticed by~\citet{Kardas2013}, existing MDD alternatives for multiagent systems have issues with the amount and quality of generated code, so they were not considered. Similarly, approaches focused on agent-based simulations (e.g., AMASON~\cite{Kluegl&Davidsson2013}, MAIA~\cite{Ghorbani+2013}) cover only basic aspects, and much is left to be manually developed. 

To achieve our goal, we derived three research questions, stated as follows. 
\begin{description}
	\item[(RQ1)] Does MDD4ABMS improve the \emph{design quality} of agent-based simulations, in comparison with NetLogo?
	\item[(RQ2)] Does MDD4ABMS decrease the \emph{effort} required to develop agent-based simulations, in comparison with NetLogo?
	\item[(RQ3)] How do developers evaluate MDD4ABMS when compared to NetLogo?
\end{description}

We selected metrics to answer these questions, described next, together with the study procedure.

\subsection{Procedure and Metrics}

The main task that was performed by our study participants consists of the development of two agent-based simulations, one in each of our target application areas (traffic and disease), hereafter referred to as domains, using one of our techniques (MDD4ABMS and NetLogo). 

Before performing this task, participants were given a hands-on training session on developing agent-based simulations.
The training session was 12-hour long. The goal is to allow participants to learn and become familiar both with the languages and tools used in the experiment (ABStractLang from MDD4ABMS, and NetLogo). The following subjects were covered in the training session, structured in the following three parts. 

\begin{description}
    \item[Part 1: Introduction to ABMS.] A background on simulations was presented to participants, and they were introduced to the ABMS paradigm. The main elements of any agent-based simulation were explained (agents, interactions, the environment, time representation, and outputs), and examples of agent-based simulations were shown. This part of the training session was completed in 1 hour. 

    \item[Part 2: Spread of disease.] The goal of this part was twofold: to provide background on spread of disease simulations, and to introduce both NetLogo and MDD4ABMS tools to the participants. The covered topics were the following, by order of presentation: agent creation and mobility, GIS files, the SIR compartmental model, and outputs for displaying populations. NetLogo was the first tool used by participants. To optimize time\footnote{A limited period of 16 hours was available for running the experiment (training + development sessions)}, the background on a particular topic was immediately followed by the NetLogo statements and functions available to implement them.   For example, after presenting the background on agent creation and mobility, participants were introduced to the NetLogo functions available for creating agents and moving them. After participants had developed a spread of disease simulation using the NetLogo functions they have learned, the topics were revisited and the ABStractLang building blocks for specifying them were shown. The same simulation was specified with ABStractlLang and have its source code generated and executed by participants. It took about 1 hour to show the basics of NetLogo and MDD4ABMS, and 5 hours to present and practice the selected topics. 
    
    \item[Part 3: Traffic signal control.] The goal of this part was to provide background on traffic signal control simulations. Considering that the participants have already become familiar with both the target tools in part 2, this part was focused on presenting only the additional NetLogo and MDD4ABMS features that are available to develop traffic signal control aspects. Consequently, this part of the training session took about 5 hours. The topics covered in this part are the following: open street map (OSM) files; how to import the specification of existing agents; traffic-related topics (e.g., phases and plans) and state machines; and the \emph{Q-learning} reinforcement learning algorithm. A set of functions that implement \emph{Q-learning} features in an existing NetLogo simulation~\cite{Roop2006} were introduced to participants and they were told to use these functions in traffic simulations developed with NetLogo. As in part 2, the background on each topic is followed by the related NetLogo statements and functions. After that, the topics were revisited and the ABStractLang building blocks were shown.
\end{description}

A three-hour slot was estimated for the participants to develop the two agent-based simulations. Each simulation had to be developed in steps, and in each of which participants were asked to develop a particular simulation feature. Details of the developed simulations are described next. 

The goal in the first simulation, referred to as \emph{Traffic}, is to evaluate how traffic signal control agents using reinforcement learning can learn a policy for selecting a traffic signal plan that minimizes the number of stopped vehicles. The features developed by participants, and associated development task, are presented next.  

\begin{description}
    \item[TF1.] \emph{Environment}: creation of a graph (traffic network) loaded from an open street map (OSM) file.
    \item[TF2.] \emph{Vehicle agent}: inclusion of the vehicle agent, following a given specification.
    \item[TF3.] \emph{Traffic signal controller agent}: creation of the traffic signal controller agent, instantiated at each intersection node of the traffic network.
    \item[TF4.] \emph{Traffic signal plans}: addition of traffic signal plans to traffic signal controllers, and set up of the coordination of the plans to manage the flow of vehicles at intersections.
    \item[TF5.] \emph{Reinforcement learning}: addition of the reinforcement learning technique (more specifically, the \emph{q-learning} algorithm) to traffic signal controllers and possibility to observe the number of stopped vehicles during the simulation.  
\end{description}

The second simulation, referred to as \emph{Disease}, aims at reproducing the perpetuation of a disease in a population of agents and has the features (with corresponding tasks) detailed next. 

\begin{description}
	\item[DF1.] \emph{Environment}: creation of a grid with a fixed size.  
	\item[DF2.] \emph{Native agent}: creation of native agents, which are positioned in the environment according to a GIS file and randomly move around the environment. 
	\item[DF3.] \emph{Disease in natives}: incorporation of the SIR compartmental model into natives, following a specification that gives the transmission, recovery, and mortality rates, as well the immunity duration and how the disease is introduced in the population of natives. Participants also had to include outputs to observe the number of susceptible, infected, and recovery natives during the simulation. Finally, agents are colored according to their current compartment (provided as coloring routines to be added to the simulation).
	\item[DF4.] \emph{Immigrant agent}: creation of the immigrant agent type, which moves exactly as natives, with a fixed population. 
	\item[DF5.] \emph{Disease in immigrants}: the SIR compartmental model is incorporated into immigrants (with specified parameters), as well as output and coloring similar to natives. Now, the disease transmission can also take place between natives and immigrants. 
\end{description}

At each step, participants were given the corresponding development task, as well as a description of the behavior they should expect when running the simulation. Therefore, participants could validate whether their MDD4ABMS model or NetLogo implementation produces the expected behavior, and they were asked to do so before finishing the step. 

The following metrics were collected while participants developed the introduced features.
\begin{description}
	\item[M1.] Number of \emph{correct features} in simulations. 
	\item[M2.] Number of \emph{incomplete features} in simulations. 
	\item[M3.] Number of \emph{features with syntactical errors} in simulations.
	\item[M4.] Number of \emph{inconsistent features} in simulations.
	\item[M5.] \emph{Time} spent to develop agent-based simulations.
\end{description}

As \citet{Hoffert+2011}, the design quality is given by the number of defects, captured by metrics M1--M4, and effort is given by the time (M5) to develop a simulation producing the expected results. The learning time is not considered in the results because during the training session both the covered topics and their exposition period were the same for both approaches.

The level at which a feature is correct is determined by inspecting the simulation model (MDD4ABMS) or implementation (NetLogo). To accomplish a particular feature and produce the expected result, a set of elements must be present in the simulation. Therefore, a feature is considered entirely correct if all its expected elements are specified by the participant. The expected constructs depend on the technique used to develop the simulation. For example, with NetLogo it is expected a set of code statements that implements all the disease-related processes (transmission, recovery, mortality, and introduction). With MDD4ABMS, in turn, a disease model capability element correctly configured is expected. As MDD4ABMS automatically generates NetLogo code from models, the size of the generated code does not affect the development effort and therefore it is not compared to the NetLogo source codes produced by participants. Such size-based effort evaluation was already considered in other work~\cite{Santos+2018simpat}. Additionally, the source code generated from MDD4ABMS models is not inspected, given that the verification procedure previously mentioned in Section~\ref{sec:mdd4abms} ensured that there is consistency between the source code and the MDD4ABMS model from which it is generated.
Given that NetLogo does not provide either traffic or disease-related features, the design quality and time metrics also capture the accuracy and development time of new functions that participants had to code in order to implement such features.

Finally, to answer RQ3, participants were asked to fill out a questionnaire (subjective evaluation), following a framework for qualitative assessment of model-driven approaches and their DSLs~\cite{Kahraman&Bilgen2015}. The following qualitative aspects were considered in this study: (i) \emph{usability}: the degree at which the approach can be used by participants to achieve their goals; (ii) \emph{reliability}: whether the approach aids producing simulations free of errors and mistakes; (iii) \emph{productivity}: the degree at which the approach promotes productivity; and (iv) \emph{expressiveness}: the degree at which it eases the development of simulations by providing elements at the right abstraction level. 

\subsection{Participants}

The study involved 31 volunteers, graduate and undergraduate students in Computer Science. Regarding demographic characteristics of participants, 90.3\% are male and 87.01\% reported age between 15--30. Participants were asked to quantify in a 9-point Likert scale their expertise in topics related to this study. Almost all participants (75\%) reported no expertise in agent-based simulations, and the remaining 25\% reported basic expertise. No participant reported expertise in NetLogo. Participants also have little expertise in topics such as traffic, epidemiology, and reinforcement learning.  Therefore, no participant needed to be excluded due to prior knowledge in the technologies under study. Our evaluation targets developers, having all of them at least basic knowledge in programming. 

Four treatment groups were adopted in the experiment to study our two independent variables, namely simulation domain (traffic or disease) and development technique (MDD4ABMS or NetLogo). Each participant was \emph{randomly} assigned to one of these groups. In each group, participants were measured in two consecutive development sessions, in which the treatment order was changed. Table~\ref{tbl:experiment-mdd4abms-treatment-group-participants} shows  the treatment conditions and the number of participants in each group. Group C has a lower number of participants: one was not considered due to a technical error (our instrumentation did not collect the time taken to perform the study tasks); and one dropped out. Given that all participants were exposed to the two techniques and two domains, the drop-out is not an indication of issues of a particular treatment.

\begin{table}
	\caption{Number of participants per treatment group.}
	\label{tbl:experiment-mdd4abms-treatment-group-participants}	
	\centering
	\footnotesize
	\begin{tabular}{ccllll} \toprule
		& & \multicolumn{2}{c}{\textbf{Development Session 1}} & \multicolumn{2}{c}{\textbf{Development Session 2}} \\ \cmidrule(l{0.5em}r{0.5em}){3-4} \cmidrule(l{0.5em}r{0.5em}){5-6}
		\textbf{Group} & \textbf{N}  & \textbf{Domain} & \textbf{Technique} & \textbf{Domain} & \textbf{Technique}  \\ \midrule
		A & 8 & Traffic & MDD4ABMS & Disease & NetLogo \\
		B & 9 & Disease & MDD4ABMS & Traffic & NetLogo \\
		C & 6 & Traffic & NetLogo & Disease & MDD4ABMS \\    
		D & 8 & Disease & NetLogo & Traffic & MDD4ABMS \\
		\bottomrule
	\end{tabular}
\end{table}

Under this presented configuration, a total of 62 simulations were developed. For each combination of domain and technique, the number of developed simulations is the following:

\begin{itemize}
\item \emph{Disease-MDD4ABMS}: 15 simulations; 
\item \emph{Disease-NetLogo}: 16 simulations; 
\item \emph{Traffic-MDD4ABMS}: 16 simulations; and 
\item \emph{Traffic-NetLogo}: 15 simulations.
\end{itemize}

\section{Results and Discussion}

The results obtained by following the described procedure are presented and discussed next. The presentation is organized into three subsections, one for each research question. Finally, we point out threats to the validity of our study and how we mitigated them.

\subsection{Design Quality (RQ1)}

We first discuss results associated with the design quality, captured by metrics M1--M4.
Table~\ref{tbl:experiment-mdd4abms-frequency-simulations-per-number-correct-features} shows the mean and standard deviation of the number of features correctly developed by participants in each simulation (M1), in addition to the number of simulations according to the number of correct features. In the traffic domain, the results obtained for MDD4ABMS and NetLogo are similar. In contrast, in the disease domain, the average number of correct features developed with MDD4ABMS (4.53) is 25\% higher than with NetLogo (3.69). In this domain, 60\% of the simulations developed in MDD4ABMS are completely correct (i.e., all 5 features correctly developed), while with NetLogo most of the simulations present either 3 (31\%) or 4 (50\%) correct features, and only 12\% (2 out of 16) are completely correct. Most of the features that were incorrectly developed in the traffic domain are the same with both MDD4ABMS and NetLogo. However, in the disease domain, participants made more mistakes while developing the transmission of the disease with NetLogo. This observation is confirmed by the significant difference among the groups that was revealed by a Kruskal-Wallis test ($\chi^2 = 14.04$, $p < 0.01$), followed by a post hoc Dunn's test with Holm correction that showed a significant difference between MDD4ABMS and NetLogo in the disease domain.

\begin{table}
\caption{Simulations and correct features.}
\label{tbl:experiment-mdd4abms-frequency-simulations-per-number-correct-features}
\footnotesize 
\centering
\begin{tabular}{@{}l@{ }r@{\hskip 0.8em}r@{\hskip 0.8em}r@{\hskip 0.8em}r@{\hskip 0.8em}r@{\hskip 0.8em}r@{}}\toprule
 \textbf{Domain /} & 	&  \multicolumn{5}{c}{\textbf{Number of simulations with N correct features}}  \\ 
 \textbf{Technique} & \textbf{Mean (SD)} 		 & \textbf{N=1} 	& \textbf{N=2} 		& \textbf{N=3}	& \textbf{N=4} 	& \textbf{N=5} \\ \midrule
  Traffic / && &&&& \\
  \hspace{0.15cm}MDD4ABMS  & 	4.56 (0.63)	&  	0 (0.00\%)  & 		0 (0.00\%)	& 	1 (6.25\%)	& 	5 (31.25\%)	&	10 (62.50\%) 	\\ 
  \hspace{0.15cm}NetLogo   & 	4.53 (0.64)	& 	0 (0.00\%)  & 		0 (0.00\%)  & 	1 (6.67\%) 	& 	5 (33.33\%)	&	9 (60.00\%)  	\\ 
  \addlinespace
  Disease / & &&&&& \\
 	\hspace{0.15cm}MDD4ABMS  & 	4.53 (0.64)	&  	0  (0.00\%) & 		0 (0.00\%)	&  	1 (6.67\%)	& 	5 (33.33\%)	&	9 (60.00\%)  	\\ 
	\hspace{0.15cm}NetLogo   & 	3.69 (0.79)	& 	0  (0.00\%) & 		1 (6.25\%)  & 	5 (31.25\%)	& 	8 (50.00\%)	&	2 (12.50\%)  	\\ % mdd increased by 25.14\%
\bottomrule
\end{tabular}
\end{table}

Features that were incorrectly developed were further analyzed, categorized according to metrics M2, M3, and M4 as follows: (M2) \emph{incomplete}, when there are missing feature elements; (M3) \emph{syntatical errors}, when there are errors that prevent running the simulation; and (M4) \emph{inconsistent}, when there is a mismatch between specification and implementation. Table~\ref{tbl:experiment-mdd4abms-incorrect-features-incomplete-synterror-inconsistent} shows the number of simulations that contain each of these defect types.

\begin{table}[h]
\caption{Number of developed simulations by defect type.}
\label{tbl:experiment-mdd4abms-incorrect-features-incomplete-synterror-inconsistent}
\footnotesize
\centering
\begin{tabular}{p{0.8cm}p{2.5cm}lrr}\toprule
 \textbf{Metric}&\textbf{Feature Defect} & \textbf{Domain} & \textbf{MDD4ABMS}& \textbf{NetLogo}  \\ \midrule
 \multirow{2}{*}{\parbox{\linewidth}{M2}}&\multirow{2}{*}{\parbox{\linewidth}{Incomplete}}  
 & Traffic & 3 (18.75\%)  &		3 (20.00\%)	 \\ 
 && Disease & 1 (6.67\%)  &	5 (31.25\%)	 	\\ 
 \midrule
 \multirow{2}{*}{\parbox{\linewidth}{M3}}&\multirow{2}{*}{\parbox{\linewidth}{Syntactical Error}}  
 & Traffic & 0 (0.00\%)  &		1 (6.67\%)	 \\ 
 && Disease & 0 (0.00\%)  &	1 (6.25\%)	 \\ 
 \midrule
 \multirow{2}{*}{\parbox{\linewidth}{M4}}&\multirow{2}{*}{\parbox{\linewidth}{Inconsistent}} 
 & Traffic & 4 (25.00\%)  &		4 (26.67\%)	 \\ 
 && Disease & 6 (40.00\%)  &	13 (81.25\%) 	\\ 
\bottomrule
\end{tabular}
\end{table}

There are incomplete features in both domains. In the traffic domain, the number of simulations with incomplete features is the same for each technique (but different percentages due to the number of participants in each group). With both techniques, there are cases in which the simulation output is missing. With MDD4ABMS, there is a simulation in which a particular traffic signal plan was not added and another with a missing relationship between one plan state machine and the reinforcement learning agent capability. With NetLogo, there are two simulations with missing code statements to activate the reinforcement learning technique.
In the disease domain, the number of simulations with incomplete features with NetLogo (31\%) is higher than with MDD4ABMS (7\%), caused by missing code statements for either initializing or managing the disease progression. With MDD4ABMS, a single simulation has a missing element (a disease immunity duration). 
There are cases of features with syntactical errors only in simulations developed with NetLogo. Errors are due to syntax issues in assignment operations and the use of agent attributes (e.g., wrong attribute names).

With respect to inconsistent features, the number of simulations is also the same in the traffic domain. With both techniques, all inconsistencies occurred while developing TF4 and TF5 (traffic signal plans and reinforcement learning, respectively). In TF4, one simulation had state machines for traffic signal plans incorrectly modeled with MDD4ABMS, while with NetLogo there are three simulations with inconsistent cycle or plan durations. In TF5, two simulations had output parameters incorrectly specified with MDD4ABMS, while two simulations have wrong learning parameters with NetLogo. 
In the disease domain, there are twice as many simulations developed with NetLogo (81\%) as with MDD4ABMS (40\%) with inconsistent features. With both techniques, most of the inconsistencies occurred while developing DF3 and DF5, in which participants had to specify the disease for the native and immigrant agents, respectively. The most common problem is the wrong specification of transmission rates for DF5. While with MDD4ABMS this inconsistency is present in 3 out of the 6 simulations with inconsistent features, with NetLogo it occurs in 11 out of the 13 simulations. This indicates that mixing programming logic with the specification of simulation parameters can possibly induce the developer to make mistakes, even when the parameters are given.

\begin{framed}
\noindent \textbf{Findings: Design Quality}. The design quality of simulations developed with MDD4ABMS is at least as good as those developed with NetLogo. In the particular case of the disease domain, the design quality is superior considering the number of correct features developed with MDD4ABMS.
\end{framed}

%%%%%%%%%%%%%%%%%%%%%%%%%%%%%%%%%
% TIME (DEVELOPMENT EFFORT)
%%%%%%%%%%%%%%%%%%%%%%%%%%%%%%%%%

\subsection{Development Effort (RQ2)}

\begin{figure}[t]
	\centering
	\includegraphics[width=0.8\linewidth]{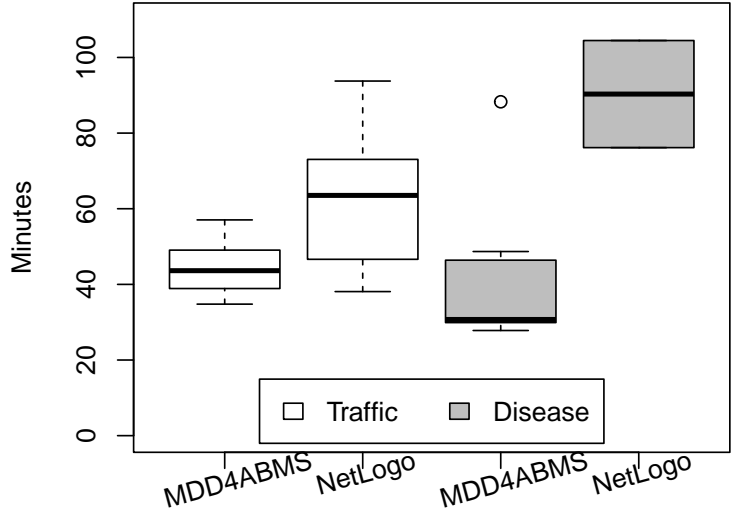} 
	%, trim={0 0 0 0.8cm},clip
	\caption{Time to develop correct simulations.}
	\label{fig:experiment-mdd4abms-time-to-develop-simulations}
\end{figure}

So far, we observed how correct the simulations developed by participants are. Now, we focus on the time taken to perform development tasks (M5). Results associated with M5 are summarized in Figure~\ref{fig:experiment-mdd4abms-time-to-develop-simulations} and detailed in Table~\ref{tbl:experiment-mdd4abms-time-to-develop-simulations}. As can be seen, participants using MDD4ABMS took less time to develop simulations in both domains.  The time taken by participants to develop the traffic simulation using MDD4ABMS (44.06 min on average) is 28\% lower than that using NetLogo (61.47 min, on average), and  55\% lower to develop the disease simulation (40.57 min with MDD4ABMS vs.\ 90.32 min with NetLogo, on average). A Kruskal-Wallis test revealed significant differences among the groups ($\chi^2 = 12.45$, $p < 0.01$), and a post hoc Dunn's test with Holm correction showed significant difference on time between MDD4ABMS and NetLogo in the disease domain.

\begin{table}
	\caption{Time (minutes) to develop correct simulations. Best mean times are boldfaced.}
	%Best mean times are highlighted in bold.
	\label{tbl:experiment-mdd4abms-time-to-develop-simulations}
	\footnotesize
	\centering
	\begin{tabular}{llrrrr}\toprule
		\textbf{Domain} & \textbf{Technique}&  \textbf{Mean} & \textbf{SD} & \textbf{Min} 	 & \textbf{Max} \\ \midrule
		\multirow{2}{*}{Traffic} 	& MDD4ABMS 	&	\textbf{44.06}	  & 	7.27 	& 		34.78	 & 	57.05		\\ % MDD is 28.32% less than NL
									& NetLogo   &	61.47	  & 	17.98   & 		38.09	 & 	93.76		\\ 
		\addlinespace			 	
		\multirow{2}{*}{Disease} 	& MDD4ABMS 	&	\textbf{40.57}	  & 	19.48	&  		27.80	 & 	88.28		\\ % MDD is 55.08% less than NL
									& NetLogo  	&	90.32	  & 	20.02   & 		76.16	 & 	104.48		\\ 
		
		\bottomrule
	\end{tabular}

\end{table}

\begin{figure}[t]
	\centering
	\subfloat[Traffic]{\includegraphics[width=0.51\linewidth]{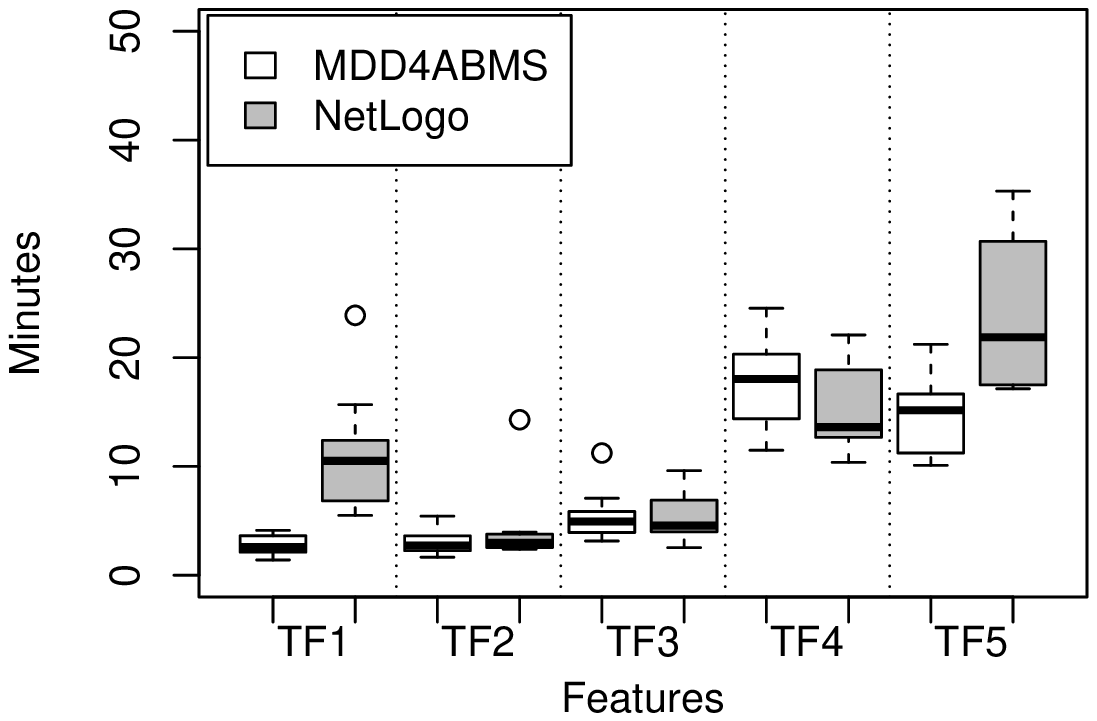}		
		\label{fig:traffic-time-to-develop-correct-simulation-features}}
	\subfloat[Disease]{\includegraphics[width=0.51\linewidth]{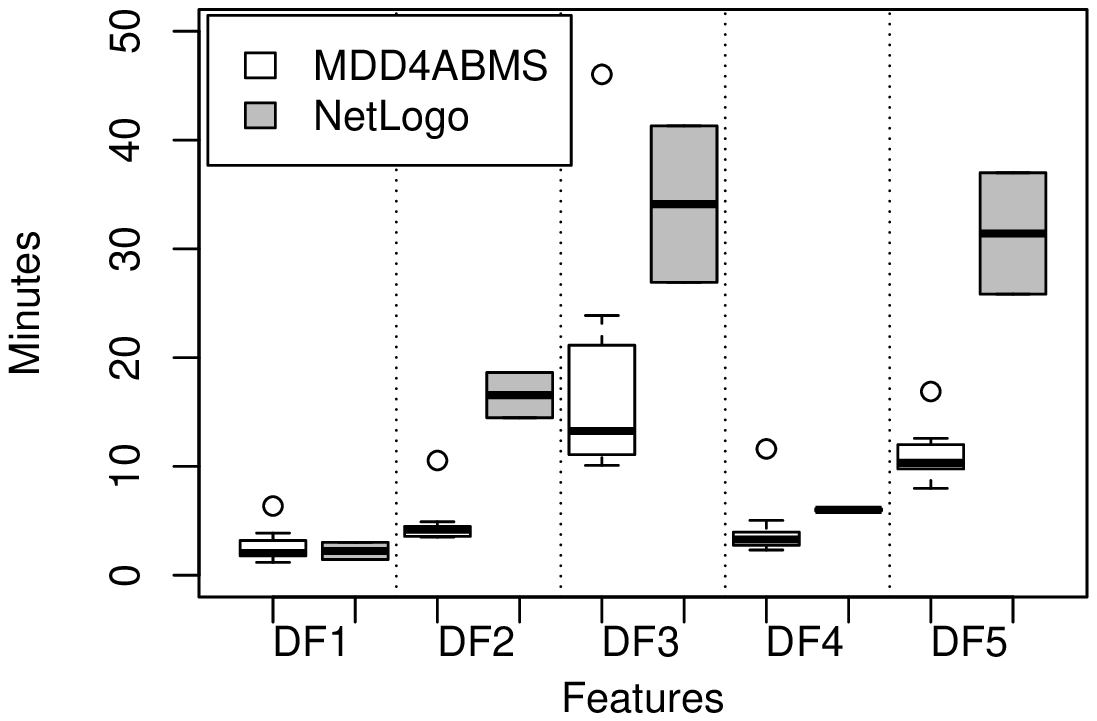}
		\label{fig:disease-time-to-develop-correct-simulation-features}}
	\caption{Time to develop correct features.}
	\label{fig:experiment-mdd4abms-time-to-develop-simulation-correct-features}
\end{figure}

Figure~\ref{fig:experiment-mdd4abms-time-to-develop-simulation-correct-features} shows the time taken by participants to develop each simulation feature. In the traffic domain (Figure~\ref{fig:traffic-time-to-develop-correct-simulation-features}), participants using MDD4ABMS took less time to develop features TF1 and TF5, and similar time for the remaining features. The more observable differences are in features that demanded sophisticated constructs. In TF1, participants had to specify a graph for the environment and initialize it from an OSM file. With NetLogo, participants had to develop data types for graph nodes and links, and write statements to open and read the file. MDD4ABMS, in contrast, provides a graph environment and participants had only to make a reference to the OSM file. In features TF2 and TF3, participants had to specify the vehicle and the traffic signal controller agent types. With both techniques, the time taken by participants to develop these features is similar. Lastly, in features TF4 and TF5, participants had to develop the traffic signal plans to control the flow of vehicles at intersections and to incorporate the reinforcement learning technique, respectively. While with NetLogo these features demand writing many lines of code, MDD4ABMS provides built-in agent capabilities for these elements. 

For feature TF4, participants took a modest higher time, on average, with MDD4ABMS in comparison with NetLogo. To investigate this unexpected result, we performed a follow-up interview with the participants. Among the respondents, many pointed out that specifying traffic signal plans with MDD4ABMS is not as straightforward as with NetLogo. They mentioned difficulties in remembering how to create and configure state machine agent capabilities, which is the means of specifying traffic signal plans. With NetLogo, traffic signal plans are implemented with quite a few lines of code to change traffic signal lights according to each plan duration. Although this ad-hoc implementation is in accordance to the feature specification, it may not be reusable in other domains. State machine agent capabilities, in MDD4ABMS, are domain-independent abstractions. The use of state machines for specifying agent behavior is reported in domains such as economics~\cite{Gnilomedov&Nikolenko2010}, manufacturing~\cite{Cicirelli+2011}, pedestrian~\cite{Sakellariou2014}, and social simulation~\cite{Adam&Gaudou2017,Adam+2017}, as well as in methodologies for specifying multiagent systems~\cite{DeLoach2014}. Additionally, some participants mentioned that due to their programming skills, it was easy to follow NetLogo examples because they easily recognized code structures (e.g.\ conditional statements and function calls).  Though state machine capabilities are abstract representations for recurrent agent-based simulation aspects, results suggest that the provided notation may not have been enough to decrease the effort to develop this aspect of traffic signal plans.

In the disease domain (Figure~\ref{fig:disease-time-to-develop-correct-simulation-features}), participants using MDD4ABMS took less time, on average, in most features. Similarly to above, the difference is more remarkable in features that demand sophisticated constructs. In feature DF1, participants only had to specify a grid environment, which was quickly developed with both techniques. In feature DF2, the number of native agents and their locations was provided by a GIS file. With NetLogo, participants had to write a couple of statements to open the file, read its content, and create the agents. With MDD4ABMS, in turn, participants just had to specify a creational strategy for the agent and refer it to the file, which took less time, on average. 
In feature DF3, while participants using NetLogo had to implement all the logic to spread the disease among agents and to recover or kill them after the infection duration, participants using MDD4ABMS had to specify a disease capability and fill it with the disease parameters. 
In feature DF4, participants had to specify the immigrant agent type and create a fixed number of agents at random locations, which was quickly developed in both techniques. 
Finally, in feature DF5 participants had to subject the immigrant agent to the disease. Once again, participants using NetLogo took more time to develop this feature because they had to implement all the transmission and recovery logic, while participants using MDD4ABMS were able to reuse the disease capability and had only to specify additional disease parameters related to the immigrant agent.  

\begin{framed}
\noindent \textbf{Findings: Development Effort}. Results indicate that MDD4ABMS decreases the development effort in comparison to NetLogo.  The effort reduction is more evident in features that require sophisticated code constructs when developed with NetLogo. In these cases, abstractions provided by MDD4ABMS reduced the development time, as the participants were able to focus on \emph{which} elements should be included in the simulation, instead of \emph{how} to implement them.
\end{framed}

%%%%%%%%%%%%%%%%%%%%%%%%%%%%%%%%%
% QUALITATIVE ASPECTS
%%%%%%%%%%%%%%%%%%%%%%%%%%%%%%%%%

\subsection{Subjective Evaluation (RQ3)}

\begin{figure}[t!]
	\centering
	\includegraphics[width=.99\linewidth,trim={1pt 0 0 0},clip]{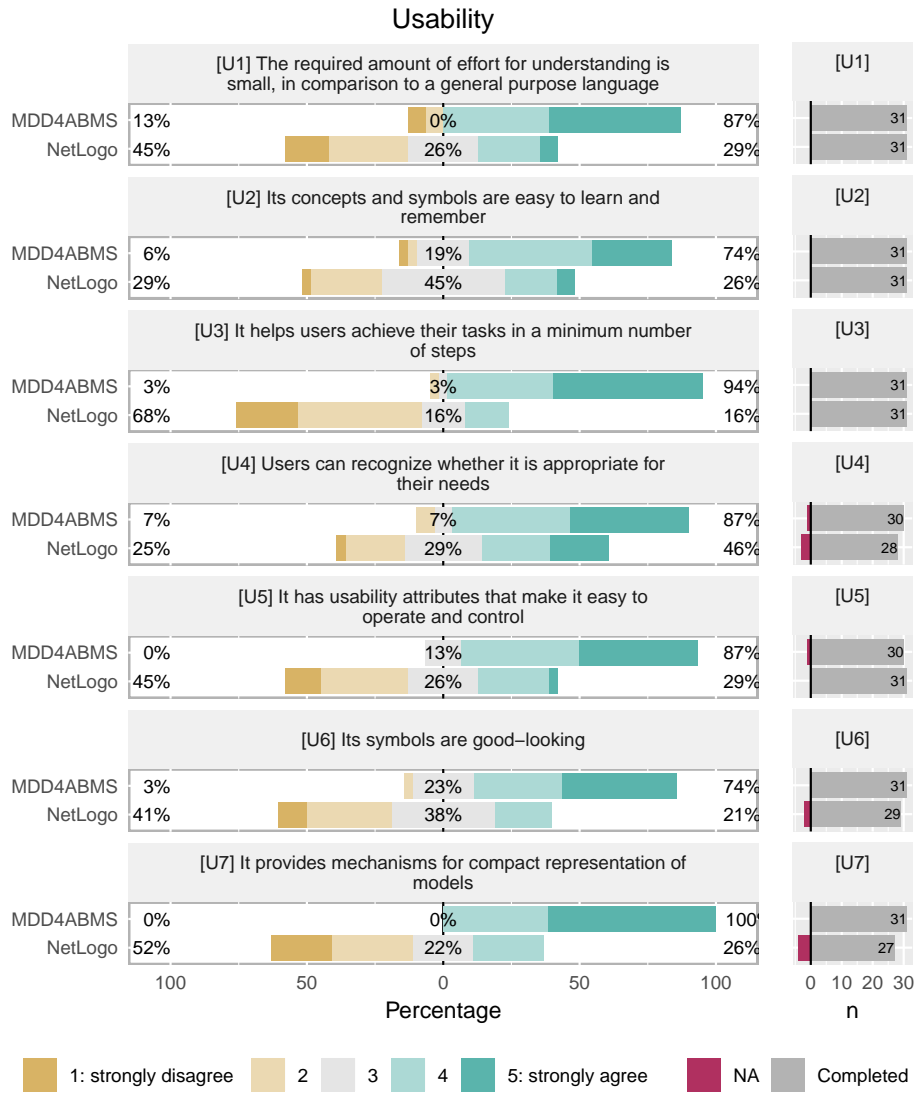}
	\caption{Subjective evaluation: usability.}
	\label{fig:experiment-mdd4abms-fqad-assessment-usability}
\end{figure}

\begin{figure}[t!]
	\centering
	\includegraphics[width=.99\linewidth,trim={1pt 1.2cm 0 0},clip]{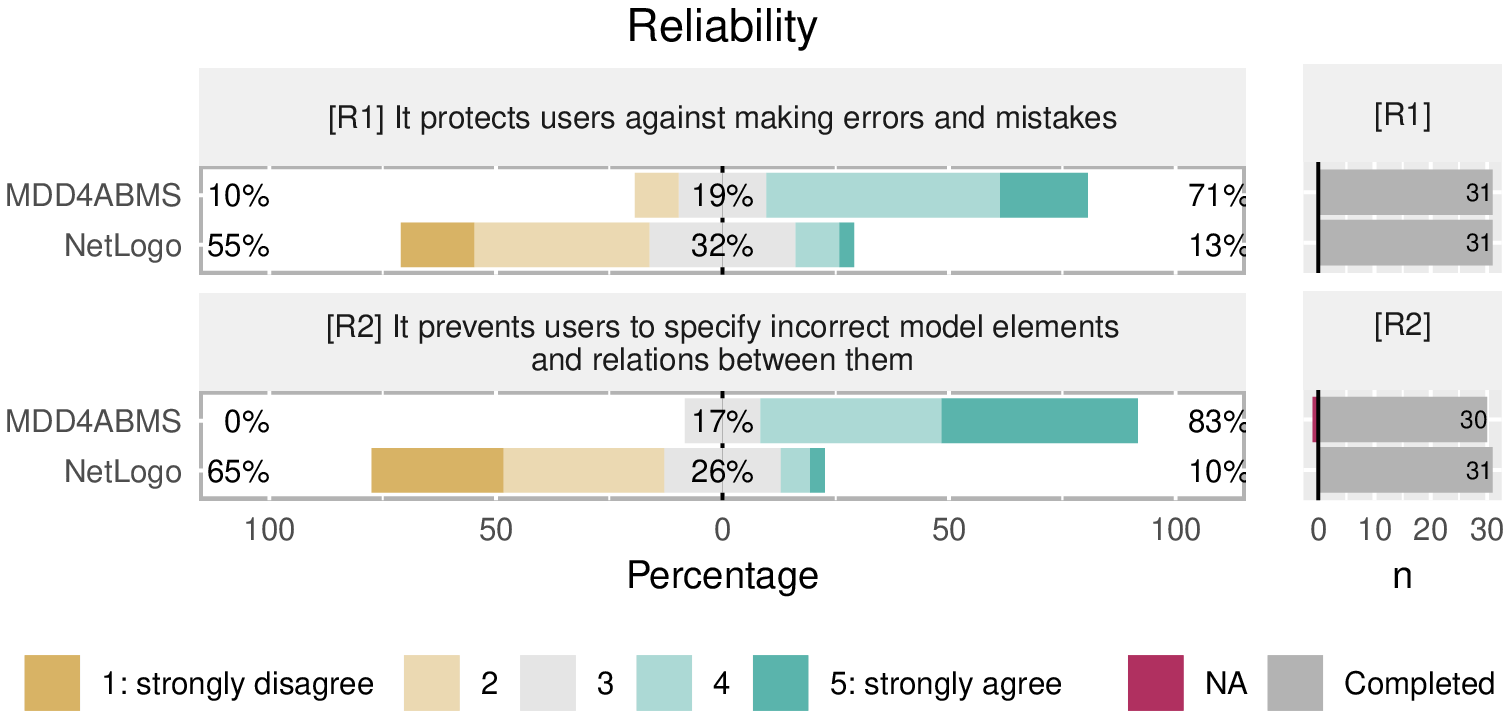}
	\includegraphics[width=.99\linewidth,trim={1pt 1.2cm 0 0},clip]{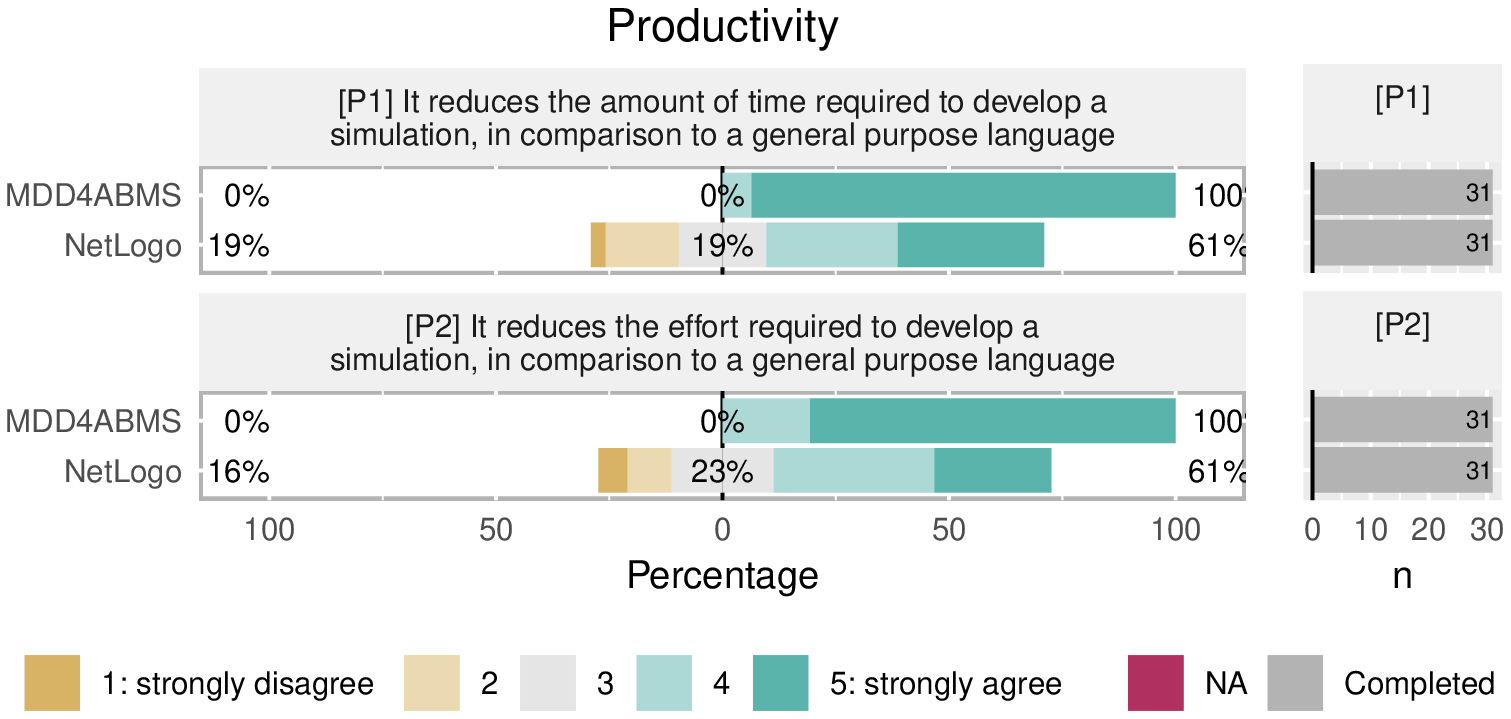}
	\includegraphics[width=.99\linewidth,trim={1pt 0 0 0},clip]{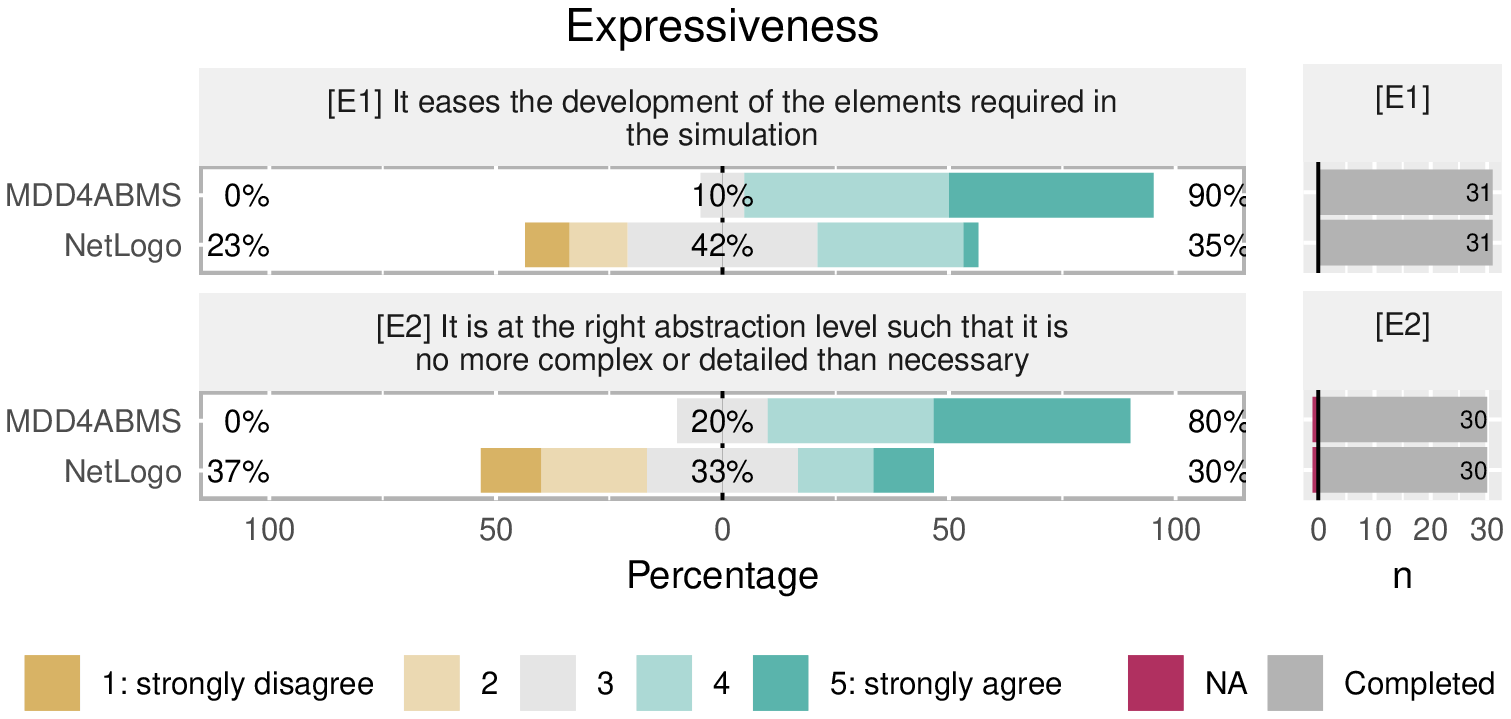}
	\caption{Subjective evaluation: reliability, productivity, and expressiveness.}
	\label{fig:experiment-mdd4abms-fqad-assessment-reliability-productivity-expressiveness}
\end{figure}

The analysis performed in the previous sections focused on objective measurements collected while participants performed the tasks of our study procedure. Participants were later requested to subjectively evaluate the two target techniques, MDD4ABMS and NetLogo, with respect to qualitative aspects. Obtained answers are summarized in Figures~\ref{fig:experiment-mdd4abms-fqad-assessment-usability} and~\ref{fig:experiment-mdd4abms-fqad-assessment-reliability-productivity-expressiveness}. The assessment was collected using the questionnaire from the qualitative assessment framework~\cite{Kahraman&Bilgen2015}, in which participants agree or disagree using a 5-point Likert scale on the presence of characteristics related to the considered qualitative aspects. In the plots of Figures~\ref{fig:experiment-mdd4abms-fqad-assessment-usability} and~\ref{fig:experiment-mdd4abms-fqad-assessment-reliability-productivity-expressiveness}, the bars indicate the level of agreement with the presented sentences, which are grouped by the aspect being assessed. Statistical tests revealed that MDD4ABSM obtained significantly higher levels of agreement across all measurements.

As can be seen, the levels of agreement with sentences associated with usability characteristics are all greater for MDD4ABMS in comparison with NetLogo. More than 80\% of participants agreed (above the intermediate level) that the effort for understanding MDD4ABMS is reduced in comparison to a general purpose language (U1), it helps users to achieve their tasks in fewer steps (U3), its users can recognize whether it is appropriate for their needs (U4), it is easy to operate (U5) and provides compact representation of models (U7)---the latter is the usability characteristic in which MDD4ABMS was best assessed: for all the participants, the level of agreement was above the intermediate level. We recall that in NetLogo modeling is done programmatically through source code, and thus compactness is related to the statements and functions provided by its language. Results also show that 74\% of the participants agreed (above the intermediate level) that MDD4ABMS concepts and symbols are easy to learn and remember (U2) and its symbols are good-looking (U6)---which according to~\citet{Kahraman&Bilgen2015} evaluates the attractiveness of the language symbols with respect to appearance and interaction. With respect to NetLogo, the levels of agreement on the presence of characteristics U3 and U7 are the lowest, meaning that participants found it laborious to develop simulations using NetLogo and that its model representation is less compact.

Similar results are observed in the other qualitative aspects. For MDD4ABMS, the levels of agreement on reliability characteristics---model checking to protect against errors (R1) and correctness (R2)---are above the intermediate level for more than 70\% of participants. In contrast, the levels of agreement for NetLogo show that, for many participants, it does not provide an effective means to avoid error making. 
The levels of agreement on the presence of productivity characteristics show that both techniques improve the development time (P1) and development effort (P2) in comparison to developing agent-based simulations using general purpose languages. However, for both of these characteristics, the levels of agreement for MDD4ABMS are greater than for NetLogo, which means that, from the point of view of participants, MDD4ABMS improves time and effort more than NetLogo (i.e. simulations are developed using MDD4ABMS in less time and with less effort), with respect to general purpose programming languages.

\begin{framed}
\noindent \textbf{Findings: Qualitative Aspects}. Results give evidence that MDD4ABMS meets qualitative aspects related to the user experience, namely usability, reliability, productivity, and expressiveness. MDD4ABMS obtained significantly higher scores with respect to all these measurements than NetLogo, when subjectively evaluated by participants. 
\end{framed}

%%%%%%%%%%%%%%%%%%%%%%%%%%%%%%%%%
% THREATS TO VALIDITY
%%%%%%%%%%%%%%%%%%%%%%%%%%%%%%%%%

\subsection{Threats to Validity}

We now present identified possible internal and external threats to validity and how we mitigated them. An internal threat is that participants could have become tired while taking part of our study, performing worse while developing the second simulation. To address this, groups have distinct treatment orders, as previously shown in Table~\ref{tbl:experiment-mdd4abms-treatment-group-participants}. This also addresses carry-over effects, given that both domain and technique were different in each development session. Another internal threat is that participants could have reused some NetLogo code developed during the tutorial sessions. To address this, participants were alerted that reuse of code would compromise the experiment results and were asked to not reuse code for a fair comparison. 
The hands-on tutorial session was given by the authors of MDD4ABMS themselves, which could have introduced a bias in the evaluation. The authors have large expertise in MDD4ABMS as well as in NetLogo (10+ years of experience). So the tutorial was not biased by the expertise in these tools. Equal rigor was adopted for teaching both tools, as well as the simulations developed during the tutorials were the same. Moreover, the tutorial included exercises so that the participants could be familiar with the approaches and their development environments before performing the study tasks. Although this may be not enough to fully master each environment, this issue affects both MDD4ABMS and NetLogo.
Another internal threat is that problems (bugs) in the ABStractLang language and its modeling tool (ABStractme) could have affected the study. To address this, tests were conducted during the development of the tool (as described in Section~\ref{sec:disease-metamodel}), and a pilot test was conducted with volunteer participants before the study (we recall that these volunteers did not participate in the main study).

An external threat to validity is that the study considered only two domains (traffic and disease), which may raise issues with respect to the generalization of results. It is important to note that while some aspects abstracted by MDD4ABMS are domain-specific (e.g., the compartmental model for the spread of diseases), others are domain-neutral (e.g., spatial abstractions, agents' creation, mobility, state machine, and reinforcement learning). 
In the presented results, we observed that in both domains there are features in which the development effort reduction was evident due to the high-level abstractions provided by MDD4ABMS. Such reduction was observed either for domain-neutral aspects---in particular, those associated with the creation of agents, initialization of the environment with external files (GIS or OSM), and reinforcement learning---and for domain-specific aspects---the spread of diseases model.  This gives evidence that the benefits provided by MDD4ABMS do not depend on the simulation domain, but on the abstraction level provided by the elements of its domain-specific modeling language. Despite the evidence, further studies can be carried out in other domains. 
As mentioned earlier in Section 4.2, the use of state machines is reported in domains such as economics and social simulation. The compartmental model can be used in other domains as well. For example, \citet{Shao&Ping2017} applied the compartmental model to simulate product diffusion in a marketing-related simulation.

Another external threat is that we only considered participants with programming expertise. Our study goal is not to generalize results for humans in general, but to developers with at least basic knowledge in programming, which matched the characteristics of our sample. However, note that programming expertise favors NetLogo and, therefore, differences could be even larger if the study was performed with participants with less expertise in programming. 

%%%%%%%%%%%%%%%%%%%%%%%%%%%%%%%%%
% FINAL REMARKS
%%%%%%%%%%%%%%%%%%%%%%%%%%%%%%%%%

\subsection{Final Remarks}
As previously described, the goal of our study is to assess the benefits of using an MDD approach to develop agent-based simulations. Participants of our study developed agent-based simulations in two distinct domains and using two techniques (MDD4ABMBS and NetLogo). The time to develop these simulations was measured, and the produced model or implementation was inspected to determine its design quality. Finally, participants subjectively assessed both MDD4ABMS and NetLogo with respect to qualitative aspects. We derived interesting findings from the study, summarized in Table~\ref{tbl:final-remarks-experiment-summary}, which also overviews details of our study design (research questions, treatments, and collected metrics).

\begin{table}[t!]
	\caption{Summary of the study and its key findings.}
	\label{tbl:final-remarks-experiment-summary}
	\footnotesize
	\centering
	\begin{tabular}{l|l}\toprule
		\multicolumn{2}{l}{\textbf{RQ1. Design Quality}}  \\ \midrule
		\textbf{Treatments} (Domain/Technique)\textbf{:} & \textbf{Findings:} \\
		Traffic / MDD4ABMS & \multirow{11}{*}[2em]{
			\noindent 
			\begin{minipage}[t]{6.5cm}
				\begin{itemize}[leftmargin=2em] 
					\item The design quality of simulations developed with MDD4ABMS is at least as good as those developed with NetLogo.
					\item In the particular case of the disease domain, the design quality is superior considering the number of correct features developed with MDD4ABMS.
				\end{itemize}
			\end{minipage}%	
		}\\
		Traffic / NetLogo & \\
		Disease / MDD4ABMS & \\
		Disease	/ NetLogo & \\ \cmidrule{1-1}
		\textbf{Metrics:} & \\ 
		M1. Correct features & \\
		M2. Incomplete features & \\
		M3. Features with syntactical errors & \\
		M4. Inconsistent features & \\
		\bottomrule
		\addlinespace
		%%%%%%%%%%%%%%%
		%%% RQ2 
		%%%%%%%%%%%%%%%
		\multicolumn{2}{l}{\textbf{RQ2. Development Effort}}  \\ \midrule
		\textbf{Treatments} (Domain/Technique)\textbf{:} & \textbf{Findings:} \\
		Traffic / MDD4ABMS & 	\multirow{8}{*}{
			\noindent 
			\begin{minipage}[t]{6.5cm}
				\begin{itemize}[leftmargin=1.5em] 
					\item MDD4ABMS decreases the development effort in comparison to NetLogo.  
					\item The effort reduction is more evident in features that require sophisticated code constructs when developed with NetLogo. In these cases, abstractions provided by MDD4ABMS reduced the development time, as the participants were able to focus on \emph{which} elements should be included in the simulation, instead of \emph{how} to implement them.
				\end{itemize}
			\end{minipage}%	
		}\\
		Traffic / NetLogo & \\
		Disease / MDD4ABMS & \\
		Disease	/ NetLogo & \\ \cmidrule{1-1}
		\textbf{Metric:} & \\ 
		M5. Time  & \\
		& \\
		& \\
		& \\
		& \\
		\bottomrule
		\addlinespace
		%%%%%%%%%%%%%%%
		%%% RQ3 
		%%%%%%%%%%%%%%%
		\multicolumn{2}{l}{\textbf{RQ3. Subjective Evaluation}}  \\ \midrule
		\textbf{Treatments:} & \textbf{Findings:} \\
		MDD4ABMS & \multirow{8}{*}[1.5em]{
			\noindent 
			\begin{minipage}[t]{6.5cm}
				\begin{itemize}[leftmargin=1.5em] 
					\item MDD4ABMS meets qualitative aspects related to the user experience. 
					\item MDD4ABMS obtained significantly higher scores with respect to all the measurements than NetLogo. 
				\end{itemize}
			\end{minipage}%	
		}\\
		NetLogo & \\ \cmidrule{1-1}
		\textbf{Qualitative aspects:} & \\ 
		Usability  & \\
		Reliability  & \\
		Productivity  & \\
		Expressiveness  & \\
		\bottomrule
	\end{tabular}	
\end{table}

%%%%%%%%%%%%%%%%%%%%%%%%%%%%%%%%%%%%%%%%%%%%%%%%%%%%%%%%%%%%%%%%%%%%%%%%%%%%%%%%%%%%%%%%%%%%%%%%%%%%%%%%%
%	CONCLUSION
%%%%%%%%%%%%%%%%%%%%%%%%%%%%%%%%%%%%%%%%%%%%%%%%%%%%%%%%%%%%%%%%%%%%%%%%%%%%%%%%%%%%%%%%%%%%%%%%%%%%%%%%%

\section{Conclusion}\label{sec:conclusion}

Model-driven approaches have been proposed to ease agent-based simulation development and increase productivity. However, existing approaches lack empirical evaluations of how they improve the development of simulations based on available platforms. Such evaluations must be conducted with humans to show concrete evidence of the benefits promoted by model-driven approaches.

In this paper, we conducted an empirical study to investigate the benefits provided by the MDD4ABMS model-driven approach, compared to the NetLogo platform, to build agent-based simulations in two application areas: adaptive traffic signal control and spread of disease. Our evaluation showed that MDD4ABMS decreases the effort to develop simulations without impacting the design quality, which is at least as good as those developed in traditional simulation platforms. These results give evidence that model-driven development is indeed a promissing alternative to ease the development and increase productivity in agent-based simulation development. Providing building blocks that simultaneously abstract recurrent simulation aspects and reduce the abstraction gap is key for that. Finally, the levels of usability, reliability, productivity, and expressiveness reported by users for MDD4ABMS are better than for traditional simulation platforms, which can foster its adoption by people with little expertise in agent-based simulation.

As future work, we will address the issue identified in our study regarding the representation of state machines in MDD4ABMS as well as extend it to be used by users with no programming expertise. The study considered novices in MDD4ABMS and NetLogo, who were given a hands-on, time-limited, training session on developing agent-based simulations. If participants had previous background on MDD4ABMS and NetLogo, it is expected that they would spend less time to develop simulations with improved design quality. Previous studies in the software industry, in which developers usually present high skills in programming languages and tools, showed that productivity is increased with MDD. Therefore, it is also expected that MDD4ABMS increases productivity in comparison to NetLogo even with skilled users. However, future studies should be conducted to identify whether this would be observed for MDD4ABMS and NetLogo skilled users. 
Finally, future studies can be conducted to evaluate the performance of the source code automatically generated by MDD4ABMS, as also to evaluate whether it is easy to correct mistakes with MDD4ABMS compared to NetLogo.

\section*{Acknowledgments}
This study was financed in part by the Coordena\c{c}\~{a}o de Aperfei\c{c}oamento de Pessoal de N\'{i}vel Superior - Brasil (CAPES) - Finance Code 001. Ingrid Nunes would like to thank CNPq for grants ref. 428157/2018-1 and 313357/2018-8.

\end{document}